\documentclass[conference]{IEEEtran}
\IEEEoverridecommandlockouts
\usepackage{cite}
\usepackage{amsmath,amssymb,amsfonts}
\usepackage{algorithmic}
\usepackage{graphicx}
\usepackage{textcomp}
\usepackage{xcolor}
\usepackage{comment}
\usepackage{url}
\usepackage{booktabs}
\usepackage{makecell} 
\usepackage{float}
\usepackage{subcaption} 

\def\BibTeX{{\rm B\kern-.05em{\sc i\kern-.025em b}\kern-.08em
    T\kern-.1667em\lower.7ex\hbox{E}\kern-.125emX}}
\begin{document}

\title{cktFormer: Transformer-Based Approach for Automated Analog Circuit Design\\
}

\author{\IEEEauthorblockN{Pasindu Dodampegama\IEEEauthorrefmark{1},
        Praveen Wijesinghe\IEEEauthorrefmark{1},
        Naveen Basnayake\IEEEauthorrefmark{1},  
        Keshawa Jayasundara\IEEEauthorrefmark{1},
        Tharindu Bandaragoda\IEEEauthorrefmark{2}
			}
	\IEEEauthorblockA{\IEEEauthorrefmark{1}University of Moratuwa, Sri Lanka, \IEEEauthorrefmark{2}Independent Researcher, Victoria, Australia\\
    \{dodampegamapd.22, wijesingherdph.22, basnayakebmns.22, jayasundaraajmkkb.22\}@uom.lk, tharindurb@gmail.com}
}

\maketitle

\begin{abstract}
Circuit design is a complex and iterative process that requires expertise in electronic engineering. It involves selecting components while meeting performance constraints, such as power efficiency, cost-effectiveness, and signal integrity. However, manual design is time-consuming and prone to errors. Although other stages of the manufacturing pipeline have benefited from AI-driven optimizations, circuit design remains a bottleneck, limiting overall productivity.  Generative AI and machine learning offer the potential to automate and improve this stage, boosting efficiency and accuracy. To address this, we introduce a dual-transformer architecture that bridges the gap between AI and circuit design by leveraging attention mechanisms to model complex, non-sequential circuit relationships. Our approach structures netlist data into graph-based representations, enabling effective learning of circuit topology and component interactions. The system consists of two interlinked models: a node prediction model that proposes components and an edge prediction model that infers valid connections. This collaborative and decoupled design captures both component-level semantics and global structural coherence. In our experiments, this architecture outperforms recent models such as AnalogGenie and cktGNN in the validity of generated circuits. By addressing key limitations in existing methods, our work advances automation in electronics engineering and contributes a benchmark for AI-driven circuit synthesis.

\end{abstract}
\section{Introduction}

In a product development cycle in an industrial environment, schematic design plays an important role and requires attention to detail and debugging. Depending on the complexity of the project, the schematic design can take weeks to months, requiring full team involvement to meet all product requirements for the product to be successful. Before designing a PCB layout, teams have to go through extensive simulations using schematic designs to confirm the feasibility and performance of the circuit. This process is time-consuming and expertise-intensive. 

\begin{figure}
    \centering
    \includegraphics[width=1.0\linewidth]{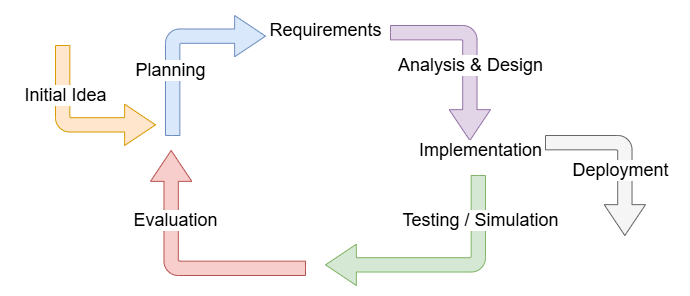}
    \caption{Model of an iterative product development process, adapted from \cite{enwiki:1259614370}.}
    \label{fig:iterative-development-model}
\end{figure}

If simulations reveal errors or unmet requirements, the team enters an iterative refinement loop (Figure~\ref{fig:iterative-development-model}): identify issues, modify the design, and re-evaluate through further simulations. If the schematic cannot meet all the requirements of the product, the team must consider their priorities and discuss possible trade-offs before the design process to find the perfect balance between feasibility, performance, cost, manufacturability, reliability and compliance with industry standards. 

During this iterative process, manual schematic design can prove to be a bottleneck, slowing development cycles and increasing the risk of time-consuming and costly errors. Many stages of electronic design have been accelerated by automation – from PCB placement/routing\cite{10028668} and VLSI chip design\cite{AMURU2023102048} to HDL code generation\cite{10.1145/3643681}, microprocessor design\cite{cheng2023pushinglimitsmachinedesign}, and analog/mixed-signal IC design\cite{article} – dramatically reducing manual effort and errors.

Despite these advances, schematic design – especially for analog circuits with continuous signals and sensitive components – remains manual and time-intensive. Our solution aims to address this issue through automation-driven design methodologies that enhance efficiency, accuracy, and scalability.

\subsection{Our Contributions}
In our work, we present a transformer-based framework for automated circuit design that addresses limitations in existing approaches, both sequential and graph-based.
Our main contributions are as follows.

\begin{itemize}
\item Component and Edge Prediction Models: We propose separate yet interlinked transformer models for node (component) and edge (connection) prediction to better capture circuit structure.

\item Encoder-Decoder transformer: We introduce a transformer-based approach that processes structured graph data by feeding both the encoder and the decoder, enabling more accurate circuit representation and generation.
\item Iterative Refinement Strategy: We implement an iterative refinement mechanism that removes any incompatible or unconnectable component after each new node is added, improving the final design’s accuracy.

\end{itemize}

Through preliminary experiments, we evaluate performance and generative capability of transformer-based models for circuit synthesis.

\section{RELATED WORK}
The broader context of circuit design automation reveals a significant divergence between digital and analog methodologies \cite{macmillen2000industrial}. Digital design automation is mature, but analog automation remains challenging due to tighter component interdependencies and less predictable behavior.

Nevertheless, with the increasing complexity of modern electronic systems and rapid advancements in Artificial Intelligence (AI), the automation of analog circuit design has gained attention in recent years. Researchers have explored frameworks ranging from sequential to graph-based learning techniques with Generative Pre-trained Transformer (GPT) models\cite{AnalogGenie}, Language Models (LMs)\cite{LaMAGIC}, and Graph Neural Networks (GNNs)\cite{hakhamaneshi2022pretraining} among other machine learning techniques.

\subsection{Sequential Approaches}
One approach to circuit design automation involves sequential representations\cite{LaMAGIC}, followed by a language model (LM)-based topology generation process. While this method achieves notable results in a single pass without requiring multiple iterations, it has limitations. Specifically, not all circuits can be efficiently represented in a strictly sequential manner, which may pose challenges in complex circuit designs.

Another innovative approach is presented in AutoCkt\cite{settaluri2020autockt}, a deep reinforcement learning framework for analog circuit design. AutoCkt mimics the sequential decision-making process of human designers by iteratively proposing design changes and then receiving feedback from a circuit simulation platform. This simulated test environment evaluates the performance of each design, allowing the model to refine its strategies until it identifies optimal solutions.

\subsection{Graph Neural Networks}

Graph Neural Networks (GNNs) \cite{GNN} have emerged as powerful tools for processing graph-structured data, making them applicable to analog circuit design. Recent work has shown that GNNs can effectively predict circuit properties \cite{hakhamaneshi2022pretraining}, reducing the reliance on costly simulations. Variants like Graph Convolutional Networks (GCNs) \cite{GCN-RL} have also been applied to tasks such as transistor sizing with promising accuracy. Using the inherent graph structure of circuits, models such as cktGNN \cite{cktGNN} use a hierarchical GNN framework and introduced the Open Circuit Benchmark (OCB) to standardize evaluations.

\subsection{Transformer Architectures}

Studies have demonstrated the effectiveness of transformers in circuit analysis, particularly in modeling nonlinear and complex circuits where traditional ML frameworks struggle \cite{AnalogandRFCircuits}. Transformers excel at capturing intricate parameter-performance relationships, making them well-suited for circuit optimization and design automation.For example, AnalogGenie employs an autoregressive transformer to generate circuit topologies. However, our dual-transformer model separates component and connection prediction, improving structure learning and circuit validity (see Table \ref{tab:valid-circuits}).

In contrast, simpler methods such as k-Nearest Neighbors (kNN) have also been evaluated in analog and RF circuit design benchmarks \cite{AnalogandRFCircuits}. kNN shows strong performance in moderately linear systems—where circuit behavior is largely predictable and data-rich—but struggles in nonlinear topologies. kNN’s local similarity focus limits its ability to model global circuit dependencies. These limitations show the need for a single model that can handle both circuit components and their connections. 

\section{METHODOLOGY}

Transformer architectures offer a compelling alternative for analog circuit synthesis by overcoming the limitations of sequential, graph-based, and instance-based models. Unlike GNN-based approaches that depend on predefined graph structures or sequential models that enforce a linear ordering, transformers use self-attention mechanisms to capture long-range dependencies and global context. This ability to model the non-sequential and interconnected nature of analog circuits makes them well-suited for this domain, allowing for expressive and flexible circuit generation.

\begin{figure}
    \centering
    \includegraphics[width=0.5\linewidth]{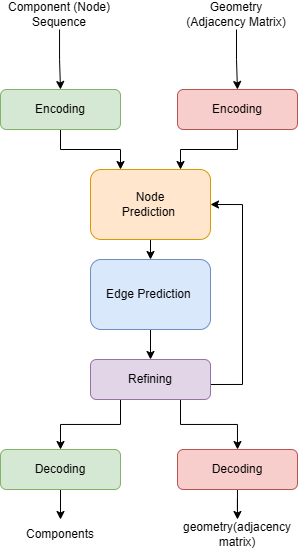}
     \caption{High-level overview of the proposed model architecture. The model first encodes circuit components and geometry, then performs edge and node prediction, followed by a refinement step to ensure circuit validity by removing or connecting unconnected nodes.}
    \label{fig:methodology}
\end{figure}

Building on these strengths, we propose a dual-transformer framework for automated analog circuit synthesis. The system comprises two interlinked models: (1) a node prediction model that sequentially proposes circuit components, and (2) an edge prediction model that infers valid electrical connections --- as illustrated in Figure \ref{fig:methodology}. Both models operate on structured graph representations of circuit schematics, enabling end-to-end generation of analog circuits that are structurally valid and functionally plausible. For all unspecified parameters and architectural components—including the scaled dot-product attention mechanism —  we adopt the standard settings and formulations from the original Transformer architecture\cite{attention_is_all}  .

\subsection{Component Encoding}
Each component of the data set is encoded in a vocabulary, as illustrated in Figure~\ref{fig:dataset}, where each unique component or terminal is assigned a distinct identifier (e.g., R1, R2). The name of a component is derived from its type (such as a resistor, capacitor, or voltage source) and an index to ensure uniqueness. Modeling component pins individually, rather than components as monolithic nodes, allows the model to learn fine-grained electrical connectivity patterns and improves edge prediction accuracy, especially in topologies with multi-terminal components like transistors. This representation provides a flexible structure for capturing electrical connectivity, enabling precise pin-to-pin modeling and simplifying edge prediction during circuit generation.

An embedding is created that maps to this vocabulary. Components are stored as embeddings in a simple lookup table, where they are mapped to numerical values. These indices are passed through a trainable embedding layer with  32 dimensions . These values are then fed into the component prediction model as decoder inputs.

Furthermore, some components like MOSFETs are broken down into separate parts — drain, source, and gate — in the vocabulary (e.g., NM1\_D → 5, NM1\_S → 6). This is because each terminal behaves differently in a circuit and plays a distinct role. Modeling these separately allows the model to reflect the MOSFET's behavior accurately in the circuit.

\subsection{Graph Encoding}
\begin{figure}
    \centering
    \includegraphics[width=0.75\linewidth]{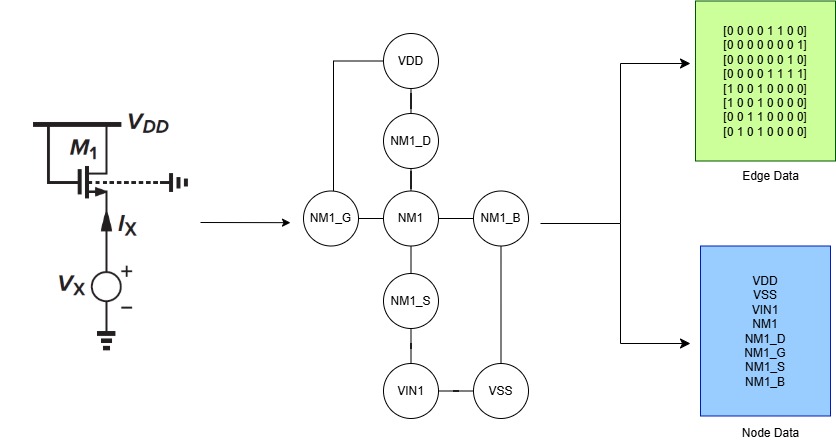}
    \caption{Encoding of a circuit schematic into component vocabulary and geometric encoding (adjacency matrix) for model input.}
    \label{fig:dataset}
\end{figure}

Graphs are encoded using a simple linear layer to match the input format required by the transformer encoder. This approach aims to represent circuit components and their interconnections in a way that is easily understandable by the model's attention mechanisms. Each electronic component/port of a component is represented as a node in the graph, with the adjacency matrix capturing the relationships between nodes. In the example provided in Figure~\ref{fig:dataset}, the first row of the adjacency matrix indicates that the V1 node is connected to Drain and Gate of the N-MOSFET (represented by the 1's). 

\subsection{Node Prediction Model}

 This model uses a Transformer-based encoder–decoder architecture. The encoder processes a fixed-dimension graph input (310 features per node) to capture structural and relational information via stacked attention layers. The decoder receives the corresponding encoded node sequence and outputs a probability distribution over a component vocabulary of size 894. This enables the model to predict the next most probable component, facilitating the addition of a new node along with its associated row and column in the adjacency matrix.

The node prediction model consists of 4 transformer layers, each with 4 attention heads. Each layer includes a multi-head graph attention block, followed by a residual connection and layer normalization. This is succeeded by a feedforward block with GeLU activation and a dropout rate of 0.1, and ending with a second residual connection and normalization. Both encoder and decoder share this structure. The model is trained independently of the edge prediction model using a cross-entropy loss and the Adam optimizer. By accepting both graph and node inputs, the model supports continuation of the circuit design from any intermediate state.

\nopagebreak

\subsection{Edge prediction Model}
The edge prediction model complements the node prediction model by determining how newly added components connect into the circuit. It uses the encoded component list as the transformer encoder input. These inputs are processed through the transformer layers, which update each component’s embedding with both local and global connection context, and finally produce connection probabilities for each pair of nodes. We develop two closely related transformer-based architectures for edge prediction that differ in input representation and internal configuration: one adopts a cyclic design, and the other a non-cyclic design.

\subsubsection{Non cyclic design}
This model takes a sequence of nodes as input and predicts possible connections in a single forward pass. Each node is represented by its embedding, with an embedding dimension of 128.  The model uses 4 transformer encoder layers (each with 8 attention heads and a feed-forward dimension of 256).

\subsubsection{Cyclic Design}
In the cyclic design, we use a Graph Transformer that incrementally predicts connections. The model is trained to take an additional incomplete adjacency matrix (used as the decoder input), along with node feature inputs, and iteratively generate the missing edges to complete the circuit. This architecture is well-suited for predicting partially designed circuits. The model uses multi-head attention with 8 heads and a 6-layer Transformer stack, with a dropout rate of 0.1 to improve generalization and prevent overfitting.

\subsubsection{Graph Attention with Edge Bias}

The cyclic model employs edge-biased graph attention to extract topological information from the circuit graph. Attention is computed between all pairs of existing nodes using their features, and edge information is integrated directly into the attention mechanism via a bias term.

To encode circuit topology, we adopt a multi-head self-attention mechanism where each attention score incorporates a learnable edge bias:
\[
\alpha_{ij}^{(h)} = \frac{\mathbf{q}_i^{(h)} \cdot \mathbf{k}_j^{(h)}}{\sqrt{d_h}} + \mathrm{EdgeBias}_{ij}^{(h)}
\]
Here, $\mathbf{q}_i^{(h)}$ and $\mathbf{k}_j^{(h)}$ are the query and key vectors for attention head $h$, and $\mathrm{EdgeBias}_{ij}^{(h)}$ is a scalar derived from the edge feature between nodes $i$ and $j$. This bias term is generated by applying a learnable linear projection to the binary adjacency matrix and is shared across all heads. This formulation allows the model to attend strongly to nodes that are connected in the circuit, guiding the attention mechanism to respect the underlying graph structure.

\subsection{Edge probability calculation}
 In both edge prediction models, attention weights are used to calculate weighted sums on the value vectors $\mathbf{v}_j$, and the outputs of all heads are concatenated and projected by the transformer to produce updated node features. Subsequently, these updated embeddings are fed into a multilayer perceptron (MLP)  comprising three linear layers interleaved with ReLU activations and a LayerNorm,  is applied in a pairwise fashion to compute connection probabilities.
\[
p_{i,\text{new}} = \sigma \left( \mathrm{MLP} \left( [\mathbf{h}_i \| \mathbf{h}_{\text{j}}] \right) \right)
\]
where \( \sigma\) is the activation of the sigmoid. This MLP outputs a scalar probability indicating the likelihood of an edge from node \(i\) to the \(j\) component. An edge is established if this probability exceeds a predefined threshold, ensuring meaningful and physically plausible connections.

\subsection{Iterative Refinement Strategy: Post-Processing and Edge Refinement}
While the prediction models provide strong initial outputs, additional steps are required to ensure physical validity and circuit consistency. In autoregressive generation, the node prediction model often proposes novel component sequences, showing generalization.  However, the subsequent edge predictions can lead to disconnected or floating nodes, indicating limitations in maintaining global structural coherence.

To improve the accuracy of edge predictions, we computed the probability for each pair of \(i,j\) nodes by
\[P_i,_j = P_j,_i=\frac{P'_i,_j+P'_j,_i}{2}\]
where \(P'\)  represents the output of the edge prediction model for each pair of nodes. This will ensure the symmetry of the generated adjacency matrix. To refine the output, we resolve conflicting and floating node connections by selecting the most probable connection. 

\subsection{Partial Circuit Completion Capability}

Previous studies, such as cktGNN\cite{cktGNN} and AnalogGenie\cite{AnalogGenie}, primarily focused on generating valid circuits under fixed constraints, with limited support for practical or functional completeness. In contrast, our model is designed to continue circuit synthesis from an intermediate point, allowing designers to collaboratively refine or extend partially completed schematics. To evaluate this capability, we used incomplete circuits generated from the test dataset and measured the accuracy of the model to predict the subsequent components and their interconnections.

\section{Experiment}
To evaluate the performance of our proposed node prediction model and edge prediction models, we conduct experiments using the dataset used in AnalogGenie\cite{AnalogGenie}.

\subsection{Dataset}

We utilize the AnalogGenie dataset as our benchmark, which includes components such as resistors, capacitors, inductors, and transistors, along with graphical and netlist representations of circuits. The dataset consists of 3351 unique analog circuits categorized by function (e.g., amplifiers, comparators, oscillators) and is split into training and testing sets in a 9:1 ratio.

\subsection{Baseline}
We compare our model with several state-of-the-art analog circuit generation methods, including CktGNN \cite{cktGNN} (VAE-based graph generation), LaMAGIC \cite{LaMAGIC} (masked graph model), AnalogCoder \cite{analogCoder} (code-generation approach), and AnalogGenie (autoregressive graph modeling). These baselines represent a diverse set of techniques, allowing for a comprehensive evaluation of our method.

\subsection{Evaluation}

\begin{figure}
    \centering
    \includegraphics[width=1\linewidth]{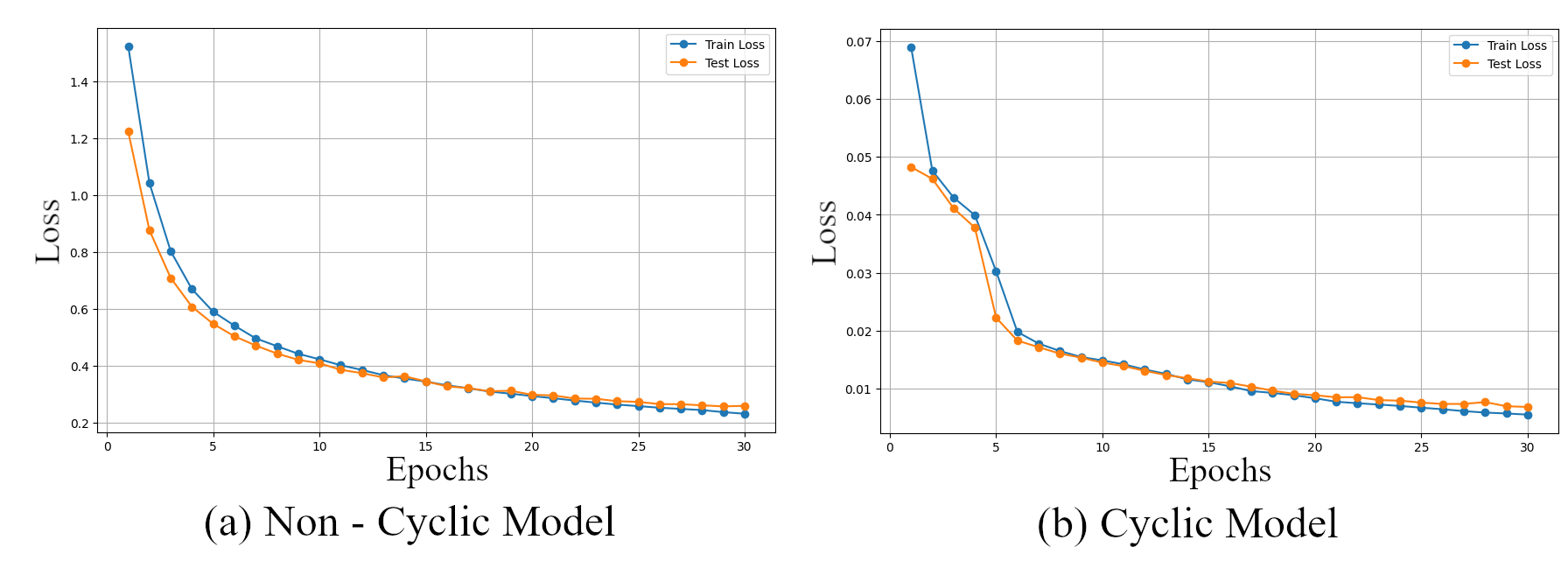}
    \caption{Training and test loss curves over epochs for edge prediction models demonstrate a consistent decrease in loss, indicating a steady training process for both models. }
    \label{fig:models}
\end{figure}

We first evaluate the graph prediction model’s learning ability, which performs well during training and validation. We use binary cross-entropy loss, calculated as:
\[
\mathcal{L}_{\text{BCE}} = -\frac{1}{N} \sum_{i=1}^{N} \left[ w_p \cdot y \cdot \log(\hat{y}) + w_n \cdot (1 - y) \cdot \log(1 - \hat{y}) 
 \right]
\]
and average it per epoch. For the cyclic graph model, we set \( w_p=w_n=1\) while for the non-cyclic model, \(w_p\) is tuned to improve accuracy and precision. As a result, the loss values differ in scale between the two graphs in Figure~\ref{fig:models}.

 Table \ref{tab:model_performance} summarizes its performance during training. The results show that the non-cyclic model  achieves the highest test accuracy and F1 score, reflecting a balanced ability to detect positive and negative edges. Its high recall makes it effective in topologies where missing an edge is costly. In contrast, the cyclic model yields the highest precision, reducing false positives, useful where incorrect connections have serious consequences.

\begin{table}
\centering
{
\begin{tabular}{|c|c|c|c|c|}
\hline
\textbf{Model} & \textbf{Test Accuracy} & \textbf{Precision} & \textbf{Recall} & \textbf{F1 Score} \\
\hline
\makecell{Non-Cyclic \\ Model} & 0.9916 & 0.7985 & 0.9226 & 0.8560 \\
\hline
\makecell{Cyclic \\ Model} & 0.9675 & 0.9310 & 0.7941 & 0.8308 \\
\hline
\end{tabular}
}
\caption{\textbf{Comparison of Edge Prediction Models:} The non-cyclic model achieves higher accuracy, recall, and F1 score, whereas the cyclic model shows higher precision.}
\label{tab:model_performance}
\end{table}

Comparison of Edge Prediction Models: The non-cyclic model achieves higher accuracy, recall and F1 score. The cyclic model achieved higher precision.

\begin{figure}
    \centering
    \includegraphics[width=1\linewidth]{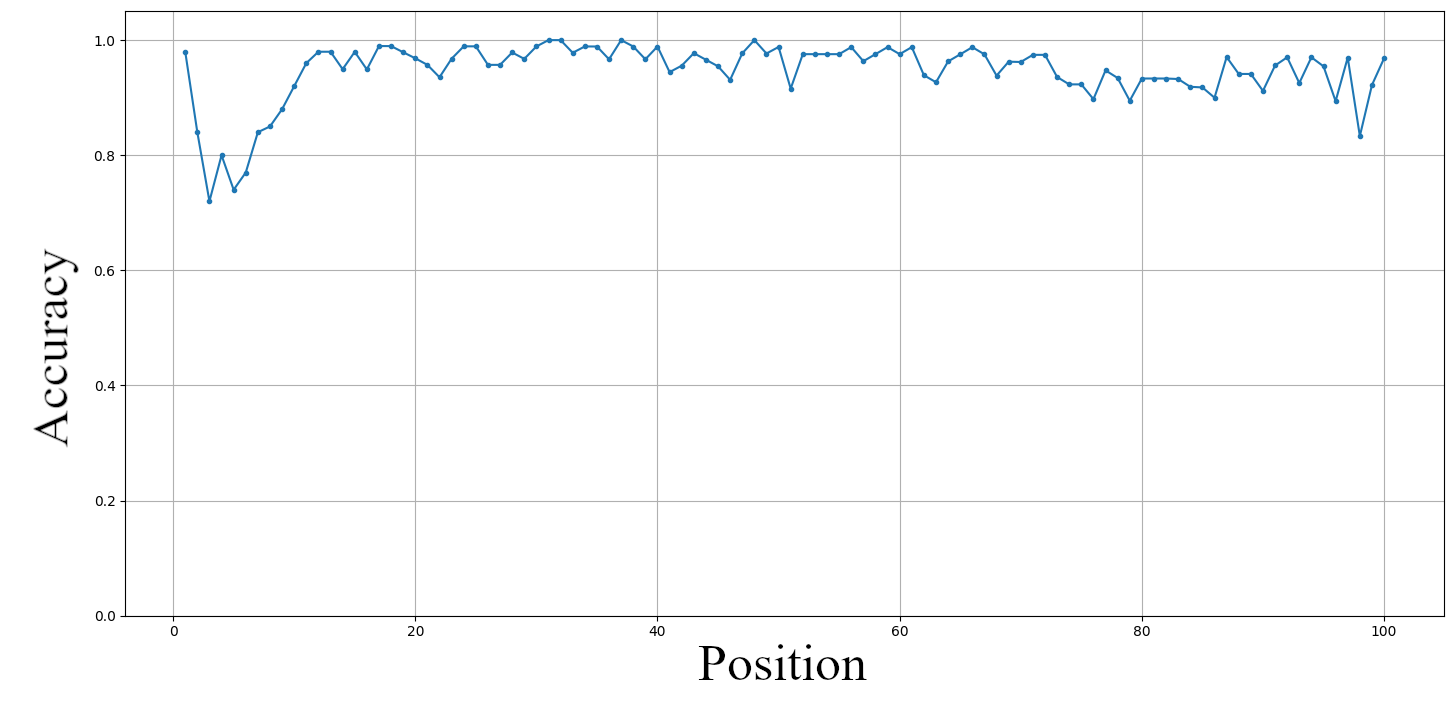}
    \caption{Node prediction accuracy by output position in the generation sequence on the test dataset. The x-axis shows output node position (number of node in input sequence + 1), and the y-axis shows average prediction accuracy, illustrating how model performance changes as nodes are added.}

    \label{fig: accuracy_by_position}
\end{figure}

Next, we evaluate the performance of the node prediction model, which is tasked with predicting the next node (i.e., the component that should be placed). The model’s average prediction accuracy was assessed in graphs with varying numbers of nodes at each positional index. The results are presented in Figure~\ref{fig: accuracy_by_position}. We observed that the node prediction accuracy is relatively low when the number of preceding nodes is fewer than 10. This suggests that the model requires a minimum context---approximately 10 nodes---to make reliable predictions. Accuracy improves as more context becomes available, indicating the model’s reliance on structural cues provided by earlier nodes. However, prediction accuracy becomes less consistent at higher positional indices. This decline is likely due to the limited large-node samples, causing unstable performance.

Transformer models typically require large data sets. In our experiments, we observed that the edge prediction model also exhibits performance limitations due to the limited size of the training dataset.

\subsection{Generation of valid circuits}
We use the following criteria—similar to those in previous works \cite{AnalogGenie}—to identify generated circuit topologies as valid and simulatable:
\begin{itemize}
\item The generated graph should be connected.
\item The node list should include a power source.
\item No short circuits should be present.
\item There should be no floating components. 
\item The power ports must be correctly connected.
\end{itemize}

We use the testing set for validation, following the same 9:1 training-to-testing ratio used in the Analog Genie paper, which demonstrates how different models generate valid circuit topologies for this dataset.

\begin{table}[h]
\centering
\begin{tabular}{lcc}
\toprule
\textbf{Evaluation Metric} & \textbf{Valid circuits (\%) $\uparrow$} \\
\midrule
CktGNN & 67.5 \\
LaMAGIC & 68.2 \\
AnalogCoder & 57.3 \\
AnalogGenie (unaug+pretrain) & 1.0 \\
AnalogGenie (aug+pretrain) & 73.5 \\

\midrule
\textbf{Cyclic Edge prediction model} & \textbf{86.4} \\
\textbf{Non-cyclic Edge prediction model} & \textbf{89.6} \\
\textbf{Non-cyclic + Node prediction model} & \textbf{76.0} \\

\bottomrule
\end{tabular}
\caption{Comparison  models based on the percentage of valid circuits generated. }

\label{tab:valid-circuits}
\end{table} 

\begin{figure}[h]
    \centering
    \includegraphics[width=1\linewidth]{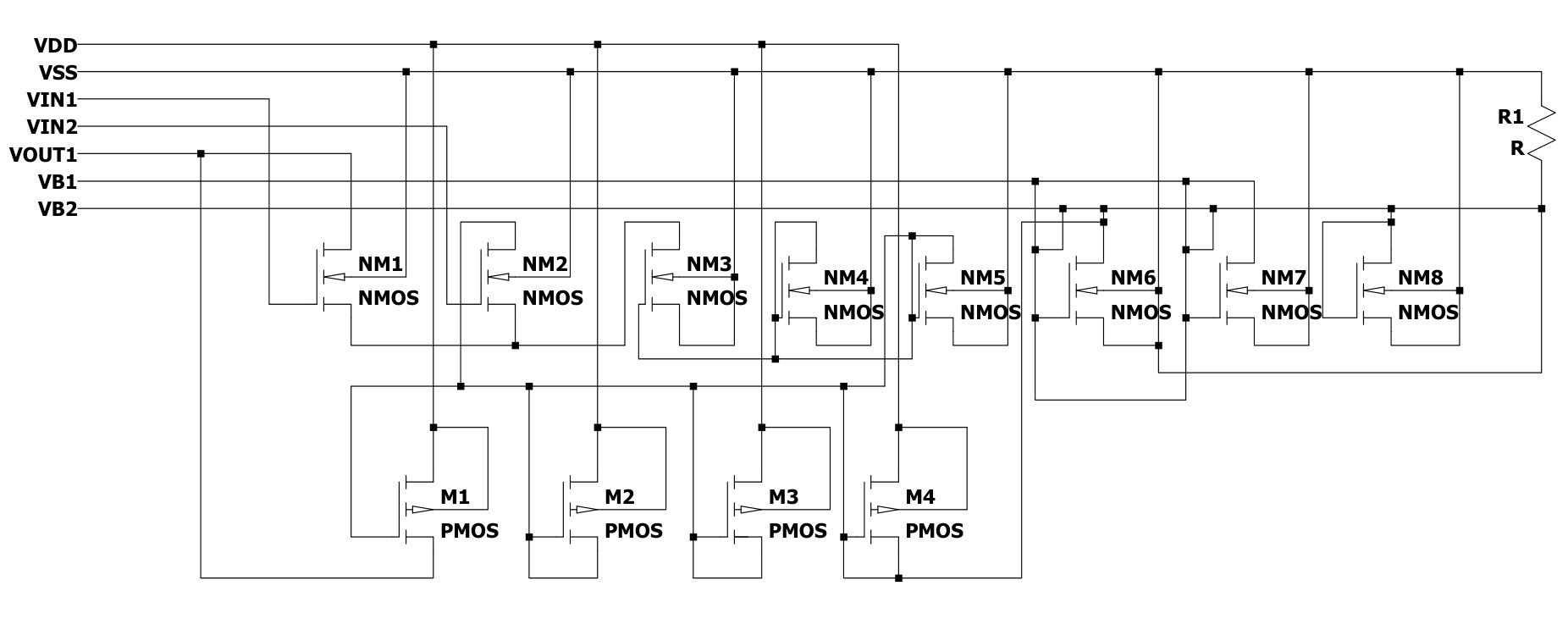}
    \caption{Example output generated by node prediction model + non-cyclic edge prediction model for an input sequence: \texttt{\{VDD, VSS, VIN1, VIN2, VOUT1, VB1, VB2, NM1, NM1\_D, NM1\_G\}}, with a randomly initialized adjacency matrix.
 }
    \label{fig:output}
\end{figure}

The generation of valid circuits is expected to be higher in edge prediction models, as their task is limited to generating edges—an inherently simpler task compared to generating an entire circuit from scratch. In contrast, the validity of circuits cannot be assessed based solely on node prediction. Therefore, the overall accuracy of a combined model is influenced by the node and edge prediction components. Our model demonstrates this expected behavior, as shown in Table \ref{tab:valid-circuits}. Circuit validity can be further improved by iterative refinement strategies that help enforce the creation of valid circuits. Figure \ref{fig:output} presents an example of a circuit generated using our model.

\section{Discussion}

Our experimental results demonstrate that transformer-based architectures offer potential for automating analog circuit design. The use of two interlinked transformer models, for node and edge prediction, addresses the non-sequential nature of circuit design by leveraging self-attention mechanisms to model complex component relationships without artificial linear constraints.

Our approach to enforcing symmetry in the adjacency matrix led to more consistent and physically realistic circuit topologies. However, some limitations remain. The model occasionally generates disconnected nodes when predicting edges for novel sequences, and its performance degrades with limited training data, highlighting the transformer architecture's reliance on large datasets. Defining rules for edge prediction can mitigate floating nodes by connecting them to the most probable node indicated by the model.

When comparing our approach to GNN-based models like cktGNN, we observe that while GNNs excel at capturing local graph structures, our transformer models better generate novel component combinations and handle partial circuit information. The higher percentage of valid circuits generated by our model (76.0\% versus 67.5\% for cktGNN) suggests that attention mechanisms may better preserve circuit validity during the generative process.

A key advantage of our approach is its ability to complete partially designed circuits, allowing it to function as a collaborative design assistant. By leveraging this capability, our model can provide autocomplete suggestions for components and connections, thereby supporting engineers throughout the design process while preserving their creative control.

\section{Conclusions and Future Work}
In this work, we propose a dual-transformer framework for automated analog circuit design, consisting of interlinked node and edge prediction models. By representing netlist data as graphs and employing self-attention mechanisms, our method captures the non-sequential and interconnected nature of analog circuits. The framework demonstrates strong performance in circuit generation tasks, and its ability to complete partially designed circuits highlights its potential as a collaborative design assistant. This study also shows the broader applicability of transformer architectures—originally developed for natural language processing—to analog circuit design.

Future models could be trained to predict not only circuit topology but also component parameters such as resistance and capacitance, improving their utility for complete designs. Integrating user-defined functional targets (e.g., gain, bandwidth) would support goal-conditioned circuit generation aligned with performance requirements. Incorporating component libraries from commercial EDA tools and training on such data can help align the system with professional workflows and industry standards. Additionally, fine-tuning for specific circuit types, such as amplifiers, could improve generation accuracy and domain-specific performance.

\section*{Acknowledgment}
The authorship team would like to acknowledge the vision, support and guidance of the IEEE Industrial Electronics Society in conducting the Generative AI Hackathon under the leadership of Daswin De Silva and Lakshitha Gunasekara.

\bibliographystyle{ieeetr}
\bibliography{citations}

@inproceedings{LaMAGIC,
author = {Chang, Chen-Chia and Shen, Yikang and Fan, Shaoze and Li, Jing and Zhang, Shun and Cao, Ningyuan and Chen, Yiran and Zhang, Xin},
title = {LaMAGIC: language-model-based topology generation for analog integrated circuits},
year = {2024},
publisher = {JMLR.org},
booktitle = {Proceedings of the 41st International Conference on Machine Learning},
articleno = {241},
numpages = {10},
location = {Vienna, Austria},
series = {ICML'24}
}

@INPROCEEDINGS{GCN-RL,
  author={Wang, Hanrui and Wang, Kuan and Yang, Jiacheng and Shen, Linxiao and Sun, Nan and Lee, Hae-Seung and Han, Song},
  booktitle={2020 57th ACM/IEEE Design Automation Conference (DAC)}, 
  title={GCN-RL Circuit Designer: Transferable Transistor Sizing with Graph Neural Networks and Reinforcement Learning}, 
  year={2020},
  volume={},
  number={},
  pages={1-6},
  doi={10.1109/DAC18072.2020.9218757}}

@inproceedings{AnalogGenie,
  author    = {Jian Gao and Weidong Cao and Junyi Yang and Xuan Zhang},
  title     = {AnalogGenie: A Generative Engine for Automatic Discovery of Analog Circuit Topologies},
  booktitle = {Proceedings of the 13th International Conference on Learning Representations (ICLR)},
  year      = {2025},
  note      = {Available: \url{https://openreview.net/forum?id=jCPak79Kev}}
}

@ARTICLE{GNN,
  author={Scarselli, Franco and Gori, Marco and Tsoi, Ah Chung and Hagenbuchner, Markus and Monfardini, Gabriele},
  journal={IEEE Transactions on Neural Networks}, 
  title={The Graph Neural Network Model}, 
  year={2009},
  volume={20},
  number={1},
  pages={61-80},
  doi={10.1109/TNN.2008.2005605}}

@misc{AnalogandRFCircuits,
      title={Supervised Learning for Analog and RF Circuit Design: Benchmarks and Comparative Insights}, 
      author={Asal Mehradfar and Xuzhe Zhao and Yue Niu and Sara Babakniya and Mahdi Alesheikh and Hamidreza Aghasi and Salman Avestimehr},
      year={2025},
      eprint={2501.11839},
      archivePrefix={arXiv},
      primaryClass={cs.LG},
      url={https://arxiv.org/abs/2501.11839}, 
      note= {Available at https://arxiv.org/abs/2501.11839}
}

@inproceedings{cktGNN,
  author    = {Zehao Dong and Weidong Cao and Muhan Zhang and Dacheng Tao and Yixin Chen and Xuan Zhang},
  title     = {CktGNN: Circuit Graph Neural Network for Electronic Design Automation},
  booktitle = {Proceedings of the 11th International Conference on Learning Representations (ICLR)},
  year      = {2023},
  note      = {Available at https://openreview.net/forum?id=NE2911Kq1sp}
}

@article{AMURU2023102048,
title = {AI/ML algorithms and applications in VLSI design and technology},
journal = {Integration},
volume = {93},
pages = {102048},
year = {2023},
issn = {0167-9260},
doi = {https://doi.org/10.1016/j.vlsi.2023.06.002},
url = {https://www.sciencedirect.com/science/article/pii/S0167926023000901},
author = {Deepthi Amuru and Andleeb Zahra and Harsha V. Vudumula and Pavan K. Cherupally and Sushanth R. Gurram and Amir Ahmad and Zia Abbas},
}

@misc{cheng2023pushinglimitsmachinedesign,
      title={Pushing the Limits of Machine Design: Automated CPU Design with AI}, 
      author={Shuyao Cheng and Pengwei Jin and Qi Guo and Zidong Du and Rui Zhang and Yunhao Tian and Xing Hu and Yongwei Zhao and Yifan Hao and Xiangtao Guan and Husheng Han and Zhengyue Zhao and Ximing Liu and Ling Li and Xishan Zhang and Yuejie Chu and Weilong Mao and Tianshi Chen and Yunji Chen},
      year={2023},
      eprint={2306.12456},
      archivePrefix={arXiv},
      primaryClass={cs.AI},
      url={https://arxiv.org/abs/2306.12456}}

@article{article,
author = {Gielen, Georges and Rutenbar, Rob},
year = {2001},
month = {01},
pages = {1825 - 1854},
title = {Computer-aided design of analog and mixed-signal integrated circuits},
volume = {88},
journal = {Proceedings of the IEEE},
doi = {10.1109/5.899053}}

@ARTICLE{10028668,
  author={Goh, Yunyeong and Jung, Dongjun and Hwang, Giyoung and Chung, Jong-Moon},
  journal={IEEE Transactions on Consumer Electronics}, 
  title={Consumer Electronics Product Manufacturing Time Reduction and Optimization Using AI-Based PCB and VLSI Circuit Designing}, 
  year={2023},
  volume={69},
  number={3},
  pages={240-249},
  doi={10.1109/TCE.2023.3240249}}

@article{10.1145/3643681,
author = {Thakur, Shailja and Ahmad, Baleegh and Pearce, Hammond and Tan, Benjamin and Dolan-Gavitt, Brendan and Karri, Ramesh and Garg, Siddharth},
title = {VeriGen: A Large Language Model for Verilog Code Generation},
year = {2024},
issue_date = {May 2024},
publisher = {Association for Computing Machinery},
address = {New York, NY, USA},
volume = {29},
number = {3},
issn = {1084-4309},
url = {https://doi.org/10.1145/3643681},
doi = {10.1145/3643681},
journal = {ACM Trans. Des. Autom. Electron. Syst.},
month = apr,
articleno = {46},
numpages = {31}
}

@misc{ enwiki:1259614370,
    author = "{Wikipedia contributors}",
    title = "Iterative and incremental development --- {Wikipedia}{,} The Free Encyclopedia",
    year = "2024",
    url = "https://en.wikipedia.org/w/index.php?title=Iterative_and_incremental_development&oldid=1259614370",
    note = "[Online; accessed 4-April-2025]"
  }

@inproceedings{attention_is_all,
author = {Vaswani, Ashish and Shazeer, Noam and Parmar, Niki and Uszkoreit, Jakob and Jones, Llion and Gomez, Aidan N. and Kaiser, \L{}ukasz and Polosukhin, Illia},
title = {Attention is all you need},
year = {2017},
isbn = {9781510860964},
publisher = {Curran Associates Inc.},
address = {Red Hook, NY, USA},
booktitle = {Proceedings of the 31st International Conference on Neural Information Processing Systems},
pages = {6000–6010},
numpages = {11},
location = {Long Beach, California, USA},
series = {NIPS'17}
}

@article{macmillen2000industrial,
  title={An industrial view of electronic design automation},
  author={MacMillen, Don and Camposano, Raul and Hill, Dwight and Williams, Thomas W},
  journal={IEEE transactions on computer-aided design of integrated circuits and systems},
  volume={19},
  number={12},
  pages={1428--1448},
  year={2000},
  publisher={IEEE}
}

@article{hakhamaneshi2022pretraining,
  title={Pretraining graph neural networks for few-shot analog circuit modeling and design},
  author={Hakhamaneshi, Kourosh and Nassar, Marcel and Phielipp, Mariano and Abbeel, Pieter and Stojanovic, Vladimir},
  journal={IEEE Transactions on Computer-Aided Design of Integrated Circuits and Systems},
  volume={42},
  number={7},
  pages={2163--2173},
  year={2022},
  publisher={IEEE}
}

@inproceedings{settaluri2020autockt,
  title={Autockt: Deep reinforcement learning of analog circuit designs},
  author={Settaluri, Keertana and Haj-Ali, Ameer and Huang, Qijing and Hakhamaneshi, Kourosh and Nikolic, Borivoje},
  booktitle={2020 Design, Automation \& Test in Europe Conference \& Exhibition (DATE)},
  pages={490--495},
  year={2020},
  organization={IEEE}
}

@inproceedings{analogCoder,
  author    = {Yuanjie Lai and Sangyun Lee and Guowei Chen and Subhajit Poddar and Mengjie Hu and David Z. Pan and Peitian Luo},
  title     = {AnalogCoder: Analog Circuit Design via Training-Free Code Generation},
  booktitle = {Proceedings of the AAAI Conference on Artificial Intelligence},
  volume    = {39},
  number    = {1},
  pages     = {379--387},
  year      = {2025},
  doi       = {10.1609/aaai.v39i1.32016},
  url       = {https://doi.org/10.1609/aaai.v39i1.32016}
}

\end{document}